\pdfoutput=1

\documentclass[11pt]{article}
\usepackage{CJKutf8}
\usepackage{float}
\usepackage{placeins} 
\usepackage{graphicx}
\usepackage{booktabs, multirow, array, makecell, caption}
\usepackage{ACL2023}
\definecolor{aqua}{rgb}{0.0, 1.0, 1.0}
\definecolor{hot pink}{rgb}{1.0, 0.31, 0.61}
\usepackage{latexsym}
\usepackage{times}  
\usepackage{helvet}  
\usepackage{courier}  
\usepackage{algorithm}
\usepackage{algorithmic}
\usepackage{tikz}
\usepackage{multirow}
\usepackage{amsmath,amssymb,mathtools,bm,etoolbox}
\usepackage{wasysym}
\usepackage{tikz}
\usepackage{xcolor}
\usepackage{soul}

\newcommand{\hlc}[2][yellow]{{%
    \colorlet{foo}{#1}%
    \sethlcolor{foo}\hl{#2}}%
}

\DeclarePairedDelimiterX{\ExpArg}[1]{[}{]}{#1}

\newcommand*\circled[1]{\tikz[baseline=(char.base)]{
            \node[shape=circle,draw,inner sep=0.3pt] (char) {#1};}}
\usepackage{newfloat}
\usepackage{listings}
\usepackage{pgfplots} 
\usetikzlibrary{patterns}
\usepackage{cancel}
\usepackage{tikz}
\newcommand{\tikzxmark}{%
\tikz[scale=0.23] {
    \draw[line width=2,line cap=round] (0,0) to [bend left=6] (1,1);
    \draw[line width=2,line cap=round] (0.2,0.95) to [bend right=3] (0.8,0.05);
}}
\newcommand{\tikzcmark}{%
\tikz[scale=0.23] {
    \draw[line width=0.7,line cap=round] (0.25,0) to [bend left=10] (1,1);
    \draw[line width=0.8,line cap=round] (0,0.35) to [bend right=1] (0.23,0);
}}
\usepackage{xcolor,pifont}
\newcommand*\colourcheck[1]{%
  \expandafter\newcommand\csname #1check\endcsname{\textcolor{#1}{\ding{52}}}%
}
\colourcheck{blue}
\colourcheck{green}
\colourcheck{red}

\newcommand{\redcross}{{\color{red}\tikzxmark}}
\newcommand{\tikzcirc}{\tikz\draw[hot pink,fill=white] (0,0) circle (0.8ex);}
\newcommand{\greentick}{{\color{blue}\tikzcmark }}
\usepackage{array}
\newcolumntype{P}[1]{>{\centering\arraybackslash}p{#1}}

\usepackage{natbib}  
\usepackage{caption} 
\usepackage[T1]{fontenc}

\usepackage[utf8]{inputenc}

\usepackage{microtype}

\usepackage{inconsolata}

\title{Advancing Multilingual Pre-training: TRIP\\Triangular Document-level Pre-training for Multilingual Language Models}

\author{
    Hongyuan Lu$^\heartsuit$, Haoyang Huang$^\spadesuit$, Shuming Ma$^\spadesuit$,\\
    \textbf{Dongdong Zhang$^\spadesuit$, Wai Lam$^\heartsuit$, Furu Wei$^\spadesuit$}\\
    $^\heartsuit$The Chinese University of Hong Kong\\
    $^\spadesuit$Microsoft Corporation\\
    \{hylu,wlam\}@se.cuhk.edu.hk\\
    \{haohua,shumma,dozhang,fuwei\}@microsoft.com
}

\begin{document}
\maketitle
\begin{abstract}
Despite the success of multilingual sequence-to-sequence pre-training, most existing approaches rely on document-level monolingual corpora in many different languages, sentence-level bilingual corpora,\footnote{In this paper, we use `bilingual corpora' to denote parallel corpora with `bilingual translation pairs' in many different language pairs, each consisting of two sentences/documents with the same meaning written in different languages. We use `trilingual corpora' to denote parallel corpora with `trilingual translation pairs' in many different language combinations, each consisting of three sentences/documents.} and sometimes synthetic document-level bilingual corpora. This hampers the performance with cross-lingual document-level tasks such as document-level 
 translation.
Therefore, we propose to mine and leverage document-level trilingual parallel corpora to improve sequence-to-sequence multilingual pre-training. We present \textbf{Tri}angular Document-level \textbf{P}re-training (\textbf{TRIP}), which is the first in the field to accelerate the conventional monolingual and bilingual objectives into a trilingual objective with a novel method called Grafting. Experiments show that TRIP achieves several strong state-of-the-art (SOTA) scores on three multilingual document-level machine translation benchmarks and one cross-lingual abstractive summarization benchmark, including consistent improvements by up to 3.11 d-BLEU points and 8.9 ROUGE-L points.
\end{abstract}

\section{Introduction}
Conventional multilingual pre-training achieved promising results on machine translation \citep{liu-etal-2020-multilingual-denoising} and cross-lingual classification \citep{xue-etal-2021-mt5}. These pre-training paradigms usually rely on monolingual corpora in many different languages, with denoising objectives such as sentence permutation and span masking \citep{liu-etal-2020-multilingual-denoising, lewis-etal-2020-bart}. Following the calls that the unsupervised scenario is not strictly realistic for cross-lingual learning \citep{artetxe-etal-2020-call}, multilingual pre-training advanced into a supervised setting through sentence-level bilingual translation pairs \citep{chi-etal-2021-mt6, reid-artetxe-2022-paradise} to provide a stronger signal for pre-training. Among these pioneering works, document-level multilingual pre-training with parallel data is currently an understudied topic. This direction is particularly significant for tasks that necessitate contextual comprehensions, such as document-level machine translation and cross-lingual summarization. As a workaround, DOCmT5 \citep{lee-etal-2022-docmt5} resorts to using synthetic bilingual translation pairs to scale up document-level multilingual pre-training.
 \par
 In addition to the lack of study for document-level multilingual pre-training with parallel data, prior works also overlooked the value of trilingual parallel data for multilingual pre-training. Compared to bilingual parallel data, trilingual parallel data is expected to better capture different linguistic clues and coherence among different languages such as past tense and gendered expressions,\footnote{For example, Chinese does not have past tense for verbs, while Japanese and English do have relevant clues. See Figure \ref{trip} for further explanation.} which can enhance the model pre-training on aspects of document-level cross-lingual understanding and resolve cross-lingual ambiguities.
\par
To this end, we present TRIP, a document-level multilingual pre-training method using trilingual parallel corpora. Because there is no publicly available document-level trilingual corpus, we propose a novel method to construct trilingual document pairs from document-level bilingual corpora. Subsequently, we augment the conventional multilingual pre-training by (i) Grafting two documents presented in two different languages into one mixed document, and (ii) predicting the remaining one language as the reference translation.
\par
We conduct experiments on document-level machine translation on TED Talks \citep{cettolo-etal-2015-iwslt}, News benchmark (News-commentary) and Europarl \citep{koehn-2005-europarl}, and cross-lingual abstractive summarization on Wikilingua \citep{ladhak-etal-2020-wikilingua,gehrmann-etal-2021-gem}. We found that TRIP clearly improves previous multilingual pre-training paradigms that use monolingual and bilingual objectives \citep{lee-etal-2022-docmt5}, and achieves strong SOTA results on both tasks.
\par
In summary, we make three key contributions:
\begin{itemize}
\setlength\itemsep{0em}
    \item TRIP proposes a novel trilingual pre-training objective through Grafting for multilingual pre-training, along with a novel method to construct trilingual data from parallel corpora.
    \item TRIP yields SOTA scores on both multilingual document-level MT and cross-lingual abstractive summarization.
    \item We conduct in-depth analyses on document-level cross-lingual understanding and compare TRIP to commercial systems.
\end{itemize}

\begin{table}[!t]
\tiny
\centering
\setlength\aboverulesep{0pt}\setlength\belowrulesep{0pt}
\setcellgapes{3pt}\makegapedcells
    \setlength\tabcolsep{2pt}
    \setlength\extrarowheight{2pt}
\begin{tabular}{l|ccccc}
\hline
\multirow{2}{*}{\textbf{Models}} & \textbf{Denoising} & \textbf{Translation}  & \textbf{Trilingual} & \textbf{Trilingual}  & \textbf{Document}\\
&\textbf{Pre-training}&\textbf{Pre-training}& \textbf{Document Pairs} & \textbf{Objective}&\textbf{Level}\\
\hline
mBART  & \greentick & \redcross & \redcross & \redcross & \greentick\\
mT5  & \tikzcirc & \redcross & \redcross & \redcross & \greentick\\
mT6  & \tikzcirc & \greentick & \redcross  & \redcross & \greentick \\
PARADISE & \tikzcirc & \greentick & \redcross & \redcross & \redcross \\
DOCmT5 & \greentick & \greentick & \redcross & \redcross & \greentick\\
\hline
\textbf{TRIP} & \greentick & \greentick & \greentick & \greentick & \greentick\\
\hline
\end{tabular}
\caption{\label{trip2}
\textbf{Comparisons of various multilingual pre-training methods.} We denote the intermediate value as \tikzcirc. For example, mT5 uses span corruption solely without sentence permutation, so we put a value of \tikzcirc \hspace{0.1mm} for the column of Denoising Pre-training for mT5. The columns of \textbf{Denoising Pre-training} and \textbf{Translation Pre-training} refer to the pre-training objectives we introduce at the start of Section \ref{S2}.
}
\end{table}

\section{Triangular Document-level Pre-training}
\label{tttrip}
\label{S2}

\begin{figure*}[t!]
\begin{center}
\centerline{
\includegraphics[width=16cm]{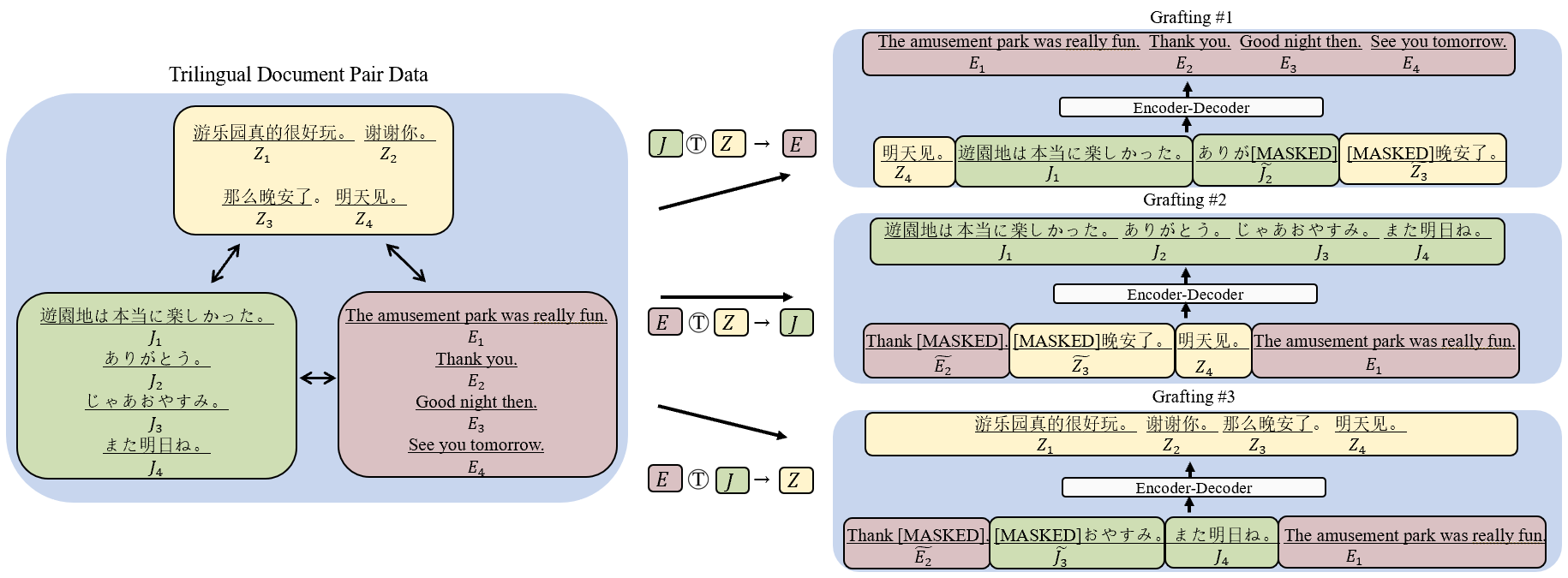}}
\caption{Overview of \textbf{Tri}angular Document-level \textbf{P}re-training (\textbf{TRIP}). We select three languages for demonstration. Here, Chinese is tenseless, and Japanese and English contain past tense as the linguistic clues that can resolve cross-lingual ambiguities. We present three Grafting cases, representing three different language combinations. For each trilingual pair, two languages serve as the input with the remaining one as the reference translation. We define a novel symbol $\circled{T}$ that denotes a noise function that combines operations in sequence: splitting by half and concatenation (\textbf{Grafting}), and sentence permutation and span corruption. $Z_n$, $J_n$, and $E_n$ for $n=\{1,2,3,4\}$ denotes four sentences written in three different languages. $\tilde{Z_n}$, $\tilde{J_n}$, and $\tilde{E_n}$ denotes corrupted sentences.
}
\label{trip}
\end{center}
\vspace{-5mm}
\end{figure*}
We start by introducing the conventional methodologies previously used by the monolingual and bilingual objectives for multilingual pre-training:
\begin{itemize}
\setlength\itemsep{0em}
    \item \textbf{Denoising Pre-training}: Sentence permutation  \citep{liu-etal-2020-multilingual-denoising} and span corruption \citep{xue-etal-2021-mt5} are effective denoising pre-training objectives for document-level multilingual pre-training.
    \item \textbf{Translation Pre-training}: Making the use of sentence-level translation pairs is a bilingual pre-training strategy for multilingual models \citep{2021arXiv210602171K,tang-etal-2021-multilingual}. 
\end{itemize}
\par
\paragraph{Constructing a Trilingual Objective}In comparison, \textbf{TRIP} is the first in the field to introduce a trilingual objective for multilingual pre-training. The core to making better use of trilingual data is to \textbf{Grafting}\footnote{Grafting refers to joining two plants together by cutting and using scion (the upper part of the grafting) as the top and the understock (the lower part of the grafting) as the root.} the documents by splitting the documents written in two different languages but with the same meaning half by half and concatenating each half to form a new document that retains the same meaning written in two different languages. TRIP then applies sentence permutation and span corruption on the Grafted documents.
\par
 Conventional monolingual and bilingual pre-training objectives overlooked the value to take such an advantage \citep{liu-etal-2020-multilingual-denoising,reid-artetxe-2022-paradise} of linguistic clues from different languages. In contrast, TRIP fuses authentic trilingual data, in which linguistic clues such as past tense and gendered nouns are usually preserved.
\par
We present in Figure \ref{trip} to illustrate how TRIP operates to make use of linguistic clues through trilingual data. Given three documents with the same meaning written in Chinese, Japanese, and English, two of the documents are split and concatenated. The concatenation is randomly permutated at the sentence level, and the remaining unchanged document is used as the translation reference. Here, Chinese is tenseless, and TRIP effectively fuses useful linguistic clues for past tense written in Japanese and English into the Chinese text to resolve cross-lingual ambiguities. 
\par

Table \ref{trip2} presents the characteristics that TRIP exhibits compared to existing methods. We report whether the models use trilingual document pairs for pre-training, and we report whether document-level tasks such as document-level machine translation or abstractive summarization are reported in their original papers. To our best knowledge, this is the first paper in our field to mine and use trilingual document pairs for multilingual pre-training. This is also the first work that features Grafting.
\par
More formally, we first denote $\mathcal{N}$ as the number of training document pairs in trilingual translation
 triplets of $(x_1, x_2, x_3)$ in a pre-training corpus $\mathcal{D}$. Given a Seq2Seq generation model \citep{S2S}, TRIP optimizes the likelihood: 
\begin{equation}
\label{eq1}
    \sum_{n=1}^{\mathcal{N}}\mathbb{E}_{x_1^n,x_2^n,x_3^n \in \mathcal{D}}[-\log P_\theta(x_3 \mid x_1  \,\,\circled{T}\,\,x_2)],
\end{equation}
where we define $\circled{T}$ as a novel operation that takes two documents in different languages as the input and takes three operations in sequence: splitting by half, concatenating, and sentence permutation.\footnote{As a pre-training method, TRIP is robust and does not require the sentences in trilingual document pairs to be perfectly aligned in their orderings. Filtering the non-perfect pairs can throw away the data and deteriorate the performance gains.}
 \paragraph{Creating Trilingual Document Pairs}
As there is no public corpus with trilingual document pairs, TRIP creates \textbf{MTDD},\footnote{Full name is masked for anonymization.} a high-quality trilingual parallel corpus with document translation pairs across 115 languages, 5,669 bilingual directions, and 99,628 trilingual combinations.
The corpus is sourced from the high-quality news documents scoped from an in-house website\footnote{Full address is masked for anonymization.} timestamped from April 2021 to July 2022. The whole procedure is composed of two steps: (i) creating bilingual document pairs and (ii) creating trilingual document pairs based on the bilingual document pairs. 
\par
To obtain bilingual document pairs, we follow ParaCrawl \citep{banon-etal-2020-paracrawl} to translate all the documents we have into English using 
 a light-weighted word-based machine translation model. The resulting translation is used for pairing only and the documents are paired and thresholded with similarity scores such as tf-idf computed on their English translation \citep{banon-etal-2020-paracrawl}. To improve efficiency, we attempt to pair documents only if they are timestamped within a small window such as one week. The motivation is that the semantic news with the same meaning  in different languages are often reported within a small timestamp window in high probabilities. The resulting document pairs are further thresholded and filtered with LASER \citep{2018arXiv181210464A},\footnote{https://github.com/facebookresearch/LASER} which is a multilingual sentence representation. 
\begin{figure}[t!]
\begin{center}
\centerline{
\includegraphics[width=7.5cm]{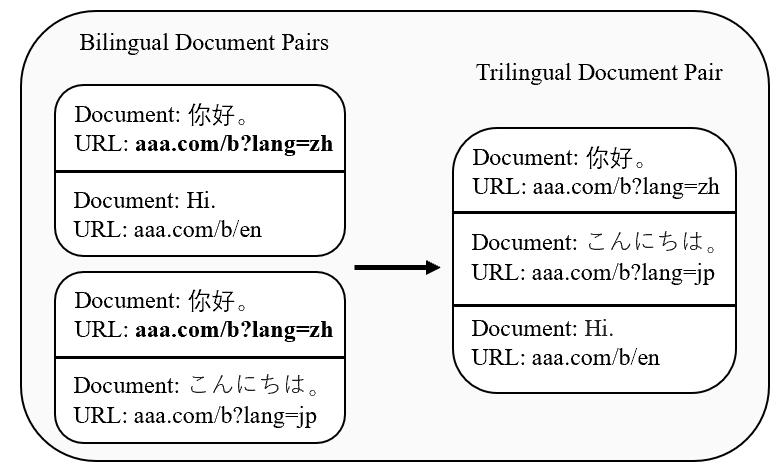}}
\caption{Illustration for the URL matching mechanism to create trilingual document pairs from bilingual data. In this case, we construct trilingual data by successfully matching the URL address for the Chinese document.
}
\label{url}
\end{center}
\vspace{-5mm}
\end{figure}
\par
Given the bilingual data constructed as above, we follow previous works \citep{banon-etal-2020-paracrawl,el-kishky-etal-2020-ccaligned}. These previous works leverage URL addresses for constructing bilingual data. In contrast, we use URL addresses to construct trilingual data pairs by matching and linking. Figure \ref{url} depicts a detailed illustration.
\par
For space reasons, we present statistics to illustrate the scale of MTDD in Table \ref{mind_lang} in Appendix \ref{mtddscale}. 
We also note that existing resources such as MTmC4 \citep{lee-etal-2022-docmt5} used by DOCmT5 can be less favourable for our experiments as (i) MTmC4 is composed of synthetic data that could be of lower quality, (ii) MTmC4 is not publicly available at the time of writing, and (iii) MTmC4 can lead to potential data leakage for the test sets on TED Talks. 

\section{Experiments}
\subsection{TRIP Pre-training}
\label{baselinemodel}
\paragraph{Model Configuration}
We use a Transformer architecture that is composed of 24 Transformer encoder layers and 12 interleaved decoder layers. In addition, it has an embedding size of 1024, and a dropout rate of 0.1. The feed-forward network is configured to have a size of 4096 with 16 attention heads. For parameter initialization, we follow  \citet{2021arXiv210613736M} and \citet{yang-etal-2021-multilingual-machine} to train a sentence-level MT system. The motivation is that previous studies have shown that the hybrid training of sentence-level and document-level MT can improve the performance of document-level translation \citep{sun-etal-2022-rethinking}. We call it the Baseline Model in the remaining of this paper.
\paragraph{Data and Pre-processing}
As described in Section \ref{S2}, we create a trilingual document-level corpus, MTDD, for TRIP pre-training with the use of trilingual document pairs. We create a list of keywords to automatically clean and remove noisy text such as claims and advertisements. We follow \citet{2021arXiv210613736M} to use SentencePiece \citep{kudo-richardson-2018-sentencepiece} for tokenization, and we use the same SentencePiece model as 
\citet{yang-etal-2021-multilingual-machine}. Following the previous works, we prefix the inputs with a language tag that indicate the target language of the generation for both pre-training and fine-tuning.
\paragraph{Training Details}
We use the Adam optimizer \citep{Adam} with $\beta_1=0.9$ and $\beta_2=0.98$ for our multilingual pre-training. The learning rate is set as $1e\text{-}5$ with a warmup step of $4000$. We use the label smoothing cross-entropy for our translation loss and we set label smoothing with a ratio of $0.1$ for model training. All of our pre-trainings are conducted on 16 NVIDIA V100 GPUs. We set the batch size as 512 tokens per GPU. To simulate a larger batch size, we update the model every 128 steps. For the Grafting operation $\circled{T}$ defined for TRIP, we split the documents 50\% by 50\%.
\begin{table}
\tiny 
\centering
\setlength\tabcolsep{2pt}
\setlength\aboverulesep{0pt}\setlength\belowrulesep{0pt}
\setcellgapes{3pt}\makegapedcells
\begin{tabular}{l|cccccc|c}
\hline
\textbf{Model} & \textbf{Fr$\rightarrow$En} & \textbf{De$\rightarrow$En} & \textbf{Zh$\rightarrow$En} & \textbf{Vi$\rightarrow$En} & \textbf{Cs$\rightarrow$En} & \textbf{Th$\rightarrow$En} & \textbf{Avg.}\\
\hline
\multicolumn{8}{c}{\textit{Sentence-level MT Models}}
\\
\hline
HAN$\dag$ & - & - & 24.00 &-&-&-&-\\
M2M-100  & 50.18 & 42.24 & 26.62 & 34.92 & 37.84 & 27.28 & 36.51\\
mBART  & 48.69 & 44.80 & 28.39 & 37.18 & 39.47 & - & -  \\
Baseline Model & 50.69 & 47.07 & 30.35 & 39.59 & 43.05 & 32.30 & 40.51\\
\hline
\multicolumn{8}{c}{\textit{Document-level MT Models}}\\
\hline
mT5$\dag$ & - & - & 24.24 &-&-&-&-\\
M2M-100  & 49.43 & 43.82 & 26.63 & 35.91 & 39.04 & 25.93 & 36.79\\
mBART  & 49.16 & 44.86 & 29.60 & 37.09 & 39.64 & - & -\\
MARGE$\dag$ & - & - & 28.40 &-&-&-&-\\
DOCmT5$\dag$ & -&-&31.40&-&-&-&-\\
Baseline Model &49.53&45.98&30.17 & 39.28 & 42.33 & 30.62 & 39.65\\
Baseline Model$^+$ & 50.74 & 46.46 & 30.65 & 39.67 & 42.64 & 31.70 &40.31\\
\hline
\textbf{TRIP (Ours)} & \textbf{51.94} & \textbf{48.24} & \textbf{31.63} & \textbf{40.52} & \textbf{44.22} & \textbf{32.87} & \textbf{41.57} \\
\hline
\end{tabular}
\caption{\label{iwslt15_en}
Results for document-level MT on TED Talks in the direction of (X $\rightarrow$ En). We report the d-BLEU scores for all the results. $\dag$: scores are taken from the official papers for these models. -: the scores are not reported or the language is not supported. The Baseline Model refers to the model described in Section \ref{baselinemodel}. The Baseline Model$^+$ represents a document-level model continually pre-trained with the bilingual data in MTDD. For a fair comparison, the trilingual data used by TRIP are constructed from these bilingual data. We perturbed them on sentence permutation and span corruption as the noise functions, with no 
 use of trilingual data.
}
\end{table}

\subsection{Multilingual Document-level MT}
\subsubsection{TED Talks}
\paragraph{Experimental Settings}
 Following DOCmT5, we use the IWSLT15 Campaign for the evaluation of TED Talks. Prior systems have reported scores on only 1 or 2 translation directions \citep{lee-etal-2022-docmt5, sun-etal-2022-rethinking}, and DOCmT5 supports only the translation direction into English (X $\rightarrow$ En). We report more language directions while DOCmT5 only evaluates on (Zh $\rightarrow$ En). Following DOCmT5, we split all documents into a maximum of 512 tokens for all train/dev/test sets during training and inference. We use the official parallel training data from IWSLT15 without any additional monolingual data, with the official 2010 dev set and 
2010-2013 test set for evaluation \citep{lee-etal-2022-docmt5}. We compute d-BLEU \citep{10.3115/1073083.1073135,liu-etal-2020-multilingual-denoising,bao-etal-2021-g}, a BLEU score for documents. We use SacreBLEU for evaluation.\footnote{\url{https://github.com/mjpost/sacrebleu}}
\paragraph{Baseline Systems}
We report strong baselines evaluated at both sentence and document levels, including SOTA models DOCmT5$\dag$ \citep{lee-etal-2022-docmt5}, M2M-100 \citep{10.5555/3546258.3546365}, mBART \citep{liu-etal-2020-multilingual-denoising}, HAN$\dag$ \citep{miculicich-etal-2018-document}, MARGE$\dag$ \citep{10.5555/3495724.3497275}, and the Baseline Model that we use to initialize the weights for TRIP. $\dag$: the scores are taken from existing papers. We also compare to the Baseline Model$^+$, a document-level model pre-trained continually on the Baseline Model with the bilingual data used to construct the trilingual data in MTDD. We do not compare to PARADISE \citep{reid-artetxe-2022-paradise}, a pre-trained model that uses dictionary denoising on monolingual data, as its weights are not publicly available so far. During our trials, we found that monolingual dictionary denoising can degrade document-level systems. We think that it could better serve sentence-level tasks such as sentence-level MT and cross-lingual classification as conducted in its original paper. See Appendix \ref{npa} for the number of model parameters.
\begin{table}
\scriptsize
\centering
    \setlength\tabcolsep{3.5pt}
\setlength\aboverulesep{0pt}\setlength\belowrulesep{0pt}
\setcellgapes{3pt}\makegapedcells
\begin{tabular}{l|cccc|c}
\hline
\textbf{Model} & \,\textbf{Fr$\rightarrow$En}\, & \,\textbf{De$\rightarrow$En}\, & \,\textbf{Zh$\rightarrow$En}\, & \,\textbf{Cs $\rightarrow$En}\, & \,\textbf{Avg.}\,\\
\hline
\multicolumn{6}{c}{\textit{Sentence-level MT Models}}
\\
\hline
M2M-100 & 31.58 & 25.65 & 18.47 & 28.17 & 25.97\\
mBART & 29.93 & 29.31 & 18.33 & 30.15 & 26.93\\
Baseline Model&35.59&34.71&27.23&37.39&33.73\\
\hline
\multicolumn{6}{c}{\textit{Document-level MT Models}}\\
\hline
M2M-100 & 32.67 & 25.78 & 17.85 & 29.06 & 26.34\\
mBART & 30.14 & 26.35 & 15.01 & 29.79 & 25.32\\
Baseline Model &36.38&34.24&25.58&36.97&33.29\\
Baseline Model$^+$ &38.47&35.20&26.74&37.26&34.42\\
\hline
\textbf{TRIP (Ours)} & \textbf{39.49} & \textbf{35.48} & \textbf{27.58} & \textbf{38.06}&\textbf{35.15}\\
\hline
\end{tabular}
\caption{\label{news_en}
Results for document-level MT on the News benchmark in the direction of (X $\rightarrow$ En).
}
\end{table}
\begin{table*}
\small 
\centering
\setlength\tabcolsep{6.25pt}
\setlength\aboverulesep{0pt}\setlength\belowrulesep{0pt}
\setcellgapes{3pt}\makegapedcells
\begin{tabular}{l|ccccccccc}
\hline
\textbf{Model} & \textbf{Da$\rightarrow$En} & \textbf{De$\rightarrow$En} & \textbf{El$\rightarrow$En} & \textbf{Es$\rightarrow$En} & \textbf{Fr$\rightarrow$En} & \textbf{It$\rightarrow$En} & \textbf{Nl$\rightarrow$En} & \textbf{Pt$\rightarrow$En} & \textbf{Sv$\rightarrow$En}\\
\hline
\multicolumn{10}{c}{\textit{Sentence-level MT Models}}
\\
\hline
M2M-100 & 50.40 & 47.38 &52.28&52.03&48.26 & 49.70 & 46.78 &49.84&52.34 \\
mBART  & - & 48.28  & - & - & 49.16 & 50.83 & 47.48 & - & -\\
Baseline Model &48.94 & 47.25 & 53.46 & 50.57 & 47.68 & 49.49 & 45.95 & 50.65 & 52.77\\
\hline
\multicolumn{10}{c}{\textit{Document-level MT Models}}\\
\hline
M2M-100 & 50.33 & 47.00 &52.24&52.14&48.13 & 49.71 & 46.65 &40.68&52.28 \\
mBART & - & 47.70 &-&-&48.98 & 50.62 & 46.96 &-&- \\
Baseline Model & 49.85& 47.64 & 53.34 & 51.32 & 48.46 & 50.26 & 47.12 & 50.13 & 52.42\\
Baseline Model$^+$ & 49.90 & 47.75 & 53.75 & 51.78 & 48.70 & 50.37 & 47.18 & 50.49 & 52.49\\
\hline
\textbf{TRIP (Ours)} & \textbf{51.13} & \textbf{48.30} & \textbf{54.38} & \textbf{52.29} & \textbf{49.36} & \textbf{51.23} & \textbf{48.07} & \textbf{51.03} & \textbf{53.43}\\
\hline
\end{tabular}
\caption{\label{europarl_main}
Results for document-level machine translation on Europarl in the direction of (X $\rightarrow$ En). 
}
\end{table*}
\paragraph{Results}
Table \ref{iwslt15_en} presents the evaluation results for TED Talks in the directions of (X $\rightarrow$ En). TRIP clearly surpasses the baselines.  TRIP surpasses the Baseline Model when both are fine-tuned at the document level by an average of 1.87 points in d-BLEU. TRIP surpasses the Baseline Model fine-tuned at the sentence level by an average of 1.01 points in d-BLEU. We postulate that the Baseline Model fine-tuned at the document level is no better than that of the sentence level due to the reason of the long input problem \citep{gtrans3}, and also due to the reason that the Baseline Model itself is pre-trained at the sentence level. TRIP beats the prior SOTA system DOCmT5. For space reasons, we present in Appendix \ref{unseenpair} the evaluations in the (X$\rightarrow$X) directions, which also show that TRIP effectively improves language pairs that are unseen during pre-training.
\par 
We also found that (i) the Baseline Model$^+$ clearly surpasses the Baseline Model and (ii) TRIP clearly surpasses the Baseline Model$^+$. This observation indicates two points: (i) the bilingual data in MTDD used to construct the trilingual data are of high quality and (ii) the trilingual objective with the Grafting mechanism is superior to the conventional bilingual objectives for multilingual pre-training.
\subsubsection{News}
\paragraph{Experimental Settings} For evaluation on the News benchmark, we follow \citet{sun-etal-2022-rethinking} to use News Commentary v11 as the training set. For Cs and De, we use newstest2015 as the dev set, and newstest2016/newstest2019 as the test set respectively. For Fr, we use newstest2013 as the dev set and newstest2015 as the test set. For Zh, we use newstest2019 as the dev set and newstest2020 as the test set. We use the same dataset preprocessing and evaluation metric as for the TED Talks.
\paragraph{Baseline Systems}
As the weights for DOCmT5 are not available at the time of writing, we compare our system to various strong baselines such as M2M-100, mBART, the Baseline Model, and the Baseline Mode$^+$. The scores are obtained by fine-tuning the official checkpoints.
\paragraph{Results} Table \ref{news_en} shows obvious and consistent improvements by up to 3.11 d-BLEU points (from 36.38 to 39.49) with TRIP for (Fr $\rightarrow$ En) compared to the Baseline Model. 
\subsubsection{Europarl}
\paragraph{Experimental Settings}
For the Europarl dataset \citep{koehn-2005-europarl}, we follow \citet{sun-etal-2022-rethinking} to use Europarl-v7, and we experiment with the setting of (X $\rightarrow$ En) where we test nine languages: Da, De, El, Es, Fr, It, Nl, Pt, and Sv. Like previous works \citep{bao-etal-2021-g, sun-etal-2022-rethinking}, the dataset is randomly partitioned into train/dev/test divisions. Additionally, we split by English document IDs to avoid information leakage.
\paragraph{Baseline Systems} As the weights for DOCmT5 are not available at the time of writing, we compare our system to various strong baselines such as M2M-100, mBART, the Baseline Model, and the Baseline Model$^+$. The scores are obtained by fine-tuning the official checkpoints.
\begin{figure}[t!]
\begin{center}
\centerline{
\includegraphics[width=7.6cm]{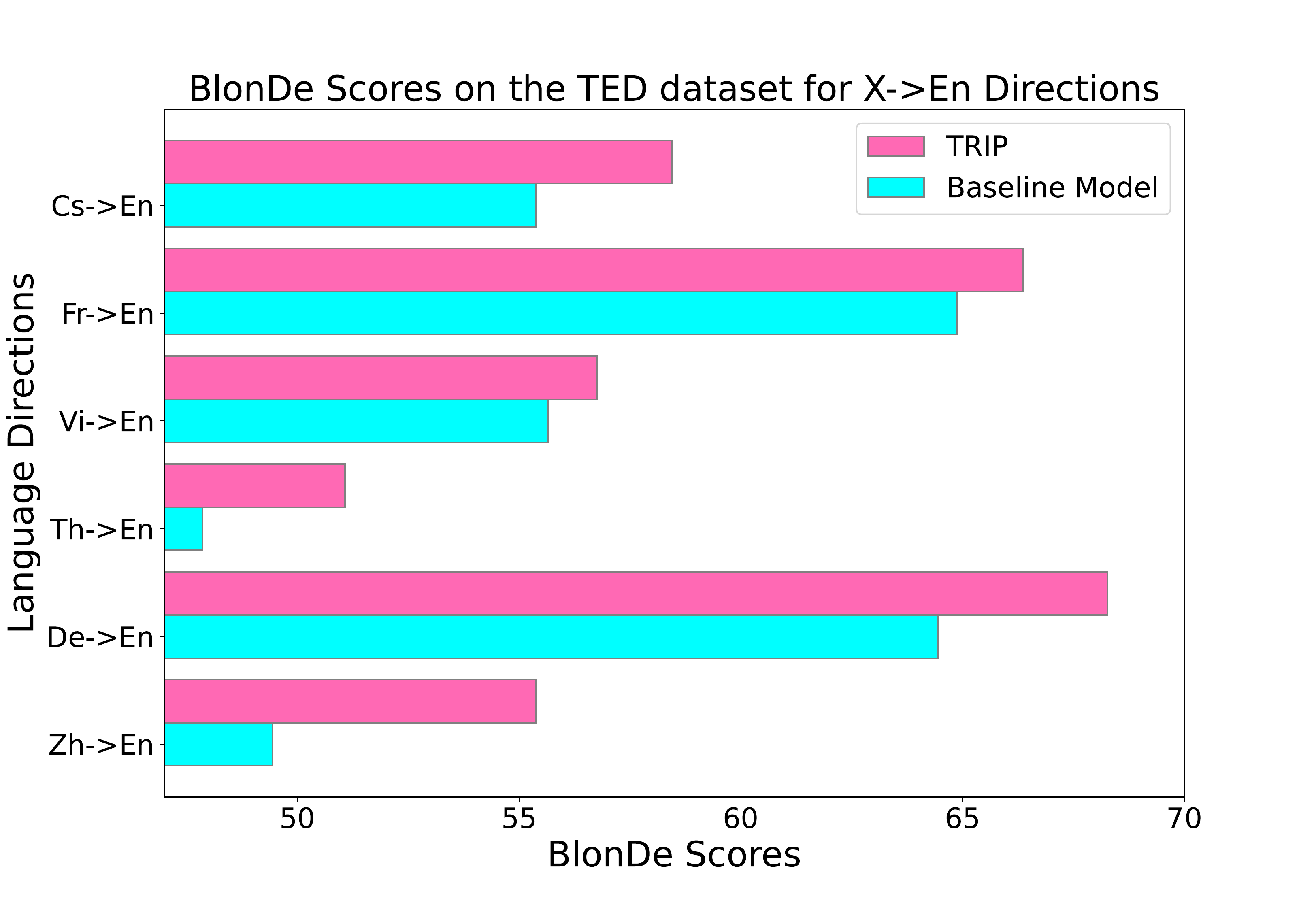}}
\caption{BlonDe scores on the TED Talks evaluated with TRIP and the Baseline Model (Document-level).
}
\label{blonde}
\end{center}
\vspace{-5mm}
\end{figure}
\begin{table*}[ht!]
\tiny 
\centering
\setlength\aboverulesep{0pt}\setlength\belowrulesep{0pt}
\setcellgapes{1.5pt}\makegapedcells
    \setlength\tabcolsep{5pt}
    \setlength\extrarowheight{1.8pt}
\begin{tabular}{l|p{11.8cm}}

\hline
\multicolumn{2}{c}{\textit{Case 1: Tense Consistency} \citep{blonde, sun-etal-2022-rethinking}}
\\
\hline
\textbf{Source} & \begin{CJK*}{UTF8}{gbsn}……， 但是\underline{这是}一个大致的抽象的讨论，当某些间隙的时候，\underline{奥克塔维奥说}，“保罗,也许我们可以观看TEDTalk。” TEDTalk用简单的方式就讲明了，……
\end{CJK*}\\
\hline
\textbf{Reference} & ..., But \hlc[aqua]{it was} a fairly abstract discussion, and at some point when there was a pause, \hlc[aqua]{Octavio said}, "Paul, maybe we could watch the TEDTalk." So the TEDTalk \hlc[aqua]{laid out} in very simple terms, ...\\
\hline
\textbf{Google Translate} & ..., But \hlc[hot pink]{it's} a roughly abstract discussion when at some point \hlc[aqua]{Octavio said}, "Paul, maybe we can watch the TEDTalk." The TEDTalk \hlc[aqua]{said} it in a simple way, ... \\
\hline
\textbf{Microsoft Translator} & ..., But \hlc[hot pink]{it's} a roughly abstract discussion when, at certain intervals, \hlc[aqua]{Octavio said}, "Paul, maybe we can watch TEDTalk." TEDTalk \hlc[hot pink]{explains} it in a simple way, ... \\
\hline
\textbf{DeepL Translate} & ..., But \hlc[aqua]{it was} a broadly abstract discussion, and when there were certain breaks, \hlc[aqua]{Octavio said}, "Paul, maybe we can watch TEDTalk." TEDTalk \hlc[hot pink]{uses}
simple way to illustrate, ... \\
\hline
\textbf{Baseline Model (Sentence-level)} & ..., But \hlc[hot pink]{it's} kind of an abstract discussion, and at some point, \hlc[hot pink]{Octavio says}, "Paul, maybe we can watch the TEDTalk." And the TEDTalk simply \hlc[hot pink]{explains} that, ... \\
\hline
\textbf{Baseline Model (Document-level)} & ..., But \hlc[hot pink]{it's} sort of an abstract discussion. And at some point, \hlc[aqua]{Octavio said}, "Paul, maybe we can watch the TEDTalk." The TEDTalk \hlc[aqua]{explained}, in a very simple way, ... \\
\hline
\hline
\textbf{TRIP} & ..., But \hlc[aqua]{it was} a sort of abstract discussion, and at some point in the intermission, \hlc[aqua]{Octavio said}, "Paul, maybe we can watch the TEDTalk." And the TEDTalk \hlc[aqua]{made} it clear, ... \\
\hline
\hline
\multicolumn{2}{c}{\textit{Case 2: Noun-related Issues} \citep{blonde}}\\
\hline
\textbf{Source} & \begin{CJK*}{UTF8}{gbsn}……，当光在\underline{西红柿}上走过时，它一直在闪耀。它并没有变暗。为什么？因为\underline{西红柿}熟了，并且光在西红柿内部反射，
……
\end{CJK*}\\
\hline
\textbf{Reference} & ..., as the light washes over \hlc[aqua]{the tomato}, It continues to glow. It doesn't become dark. Why is that? Because \hlc[aqua]{the tomato is actually ripe}, and  \hlc[aqua]{the light} is bouncing around inside the tomato, ...\\
\hline
\textbf{Google Translate} & ..., as the light passed over \hlc[hot pink]{the tomatoes}, It kept shining. It didn't get darker. Why? Because \hlc[hot pink]{the tomatoes are ripe}, and \hlc[hot pink]{light} is reflected inside the tomatoes, ... \\
\hline
\textbf{Microsoft Translator} & ..., as the light walks over \hlc[hot pink]{the tomatoes}, It keeps shining. It didn't darken. Why? Because \hlc[hot pink]{the tomatoes are ripe}, and \hlc[hot pink]{light} is reflected inside the tomatoes, ... \\
\hline
\textbf{DeepL Translate} & ..., as the light traveled over \hlc[hot pink]{the tomatoes}, it kept shining. It doesn't dim. Why? Because \hlc[hot pink]{the tomatoes are ripe} and \hlc[aqua]{the light} is reflecting inside the tomatoes, ...\\
\hline
\textbf{Baseline Model (Sentence-level)} & ..., as the light goes over \hlc[aqua]{the tomato}, It's always glowing. It's not darkening. Why? Because \hlc[aqua]{the tomato is ripe}, and  \hlc[hot pink]{light} is reflected inside the tomato, ... \\
\hline
\textbf{Baseline Model (Document-level)} & ..., as the light passes over \hlc[aqua]{the tomato}, It keeps flashing. It doesn't get darker. Why? Because \hlc[hot pink]{the tomatoes are ripe} , and  \hlc[aqua]{the light} is is reflected inside the tomato, ... \\

\hline
\hline
\textbf{TRIP} & ..., as the light passes over \hlc[aqua]{the tomato}, It's flashing all the time. It's not getting darker. Why? Because \hlc[aqua]{the tomato is ripe}, and \hlc[aqua]{the light} is reflected inside the tomato, ... \\
\hline
\multicolumn{2}{c}{\textit{Case 3: Conjunction Presence} \citep{10.1609/aaai.v33i01.33017338, sun-etal-2022-rethinking}}\\
\hline
\textbf{Source} & \begin{CJK*}{UTF8}{gbsn}……， 
我想提醒大家，我已经谈论了很多前人的事情。\underline{我还想考虑一下}，民主会是什么样子,或者是已经是什么样子的可能性 如果我们可以让更多的母亲参与进来，
……
\end{CJK*}\\
\hline
\textbf{Reference} & ..., I want to suggest to you that I've been talking a lot about the fathers. \hlc[aqua]{And I want to think} about the possibilities of what democracy might look like, or might have looked like, if we had more involved the mothers, ...\\
\hline
\textbf{Google Translate} & ..., I want to remind everyone that I've talked a lot about my predecessors. \hlc[hot pink]{I also want to think} about what democracy would look like, or is it already
What the possibilities look like if we could get more mothers involved, ... \\
\hline
\textbf{Microsoft Translator} & ..., I want to remind you that I have talked a lot about my predecessors. \hlc[hot pink]{I would also like to consider} what democracy would look like, or already be
What kind of possibilities if we can involve more mothers
, ... \\
\hline
\textbf{DeepL Translate} & ..., I want to remind you that I've talked about a lot of things that have come before.  \hlc[hot pink]{I also want to consider} the possibility of what democracy would look like, or what it already looks like if we could get more mothers involved in, ...\\
\hline
\textbf{Baseline Model (Sentence-level)} & ..., I want to remind you that I've talked about a lot of my predecessors. \hlc[hot pink]{I also want to think} about what democracy might look like, or what democracy might look like if we could get more mothers involved, ... \\
\hline
\textbf{Baseline Model (Document-level)} & ..., I'd like to remind you that I've talked about a lot of things before. \hlc[hot pink]{I'd also like to think} about the possibilities of what democracy might look like, or what it might be like, if we could get more mothers to participate, ... \\
\hline
\hline
\textbf{TRIP} & ..., I want to remind you that I've talked a lot about the past. \hlc[aqua]{And I want to think} about the possibilities of what democracy might look like, or already looks like, if we can get more mothers involved, ... \\
\hline
\end{tabular}
\caption{\label{case}
Three case studies from TED Talks demonstrate that TRIP captures better tense consistency, noun-related issues, and conjunction presence. We highlight the correct translation in aqua (darker one when printed in B\&W), and the mistakes in hot pink (lighter one when printed in B\&W). Google Translate: \url{https://translate.google.com/}, Microsoft Translator: \url{https://www.bing.com/translator}, DeepL Translate: \url{https://www.deepl.com/translator}.
}
\end{table*}

\paragraph{Results} By comparing TRIP to strong baselines, we see that the improvements with TRIP are consistent in all directions, and surpass all the strong baselines. This validates TRIP's effectiveness.
\subsubsection{Coherence and Consistency Evaluation}
\paragraph{BlonDe Evaluation} Figure \ref{blonde} depicts the evaluations on TED Talks with BlonDe scores \citep{blonde}, an evaluation metric designed for document-level MT which considers coherence and consistency issues that require the model to resolve cross-lingual ambiguities. Consistent improvements can be observed in all the directions on TED Talks with TRIP, meaning that TRIP generates more coherent and consistent translations than the baseline does. As discussed in Section \ref{tttrip}, we postulate that these improvements attribute to the Grafting mechanism that resolves cross-lingual ambiguities by exploiting useful linguistic clues in trilingual data. This improves translation in coherence and consistency as reflected in the BlonDe scores. We demonstrate case studies for more analysis of coherence and consistency issues.
\begin{table*}
\scriptsize 
\centering
\setlength\tabcolsep{7pt}
\setlength\aboverulesep{0pt}\setlength\belowrulesep{0pt}
\setcellgapes{2.5pt}\makegapedcells
\begin{tabular}{l|ccccccc}
\hline
\textbf{Model} &  \textbf{Tr$\rightarrow$En} & \textbf{Vi$\rightarrow$En} & \textbf{Ru$\rightarrow$En} &
\textbf{Es$\rightarrow$En} & \textbf{Hi$\rightarrow$En} & \textbf{Fr$\rightarrow$En} & \textbf{Id$\rightarrow$En}\\
\hline
mT5-XL$\dag$ & 40.0/18.3/33.3
& 37.6/14.9/31.2
& \textbf{37.2/14.6/30.9}
& \textbf{41.2/17.2/34.6} & -/-/- & -/-/-  & -/-/-
\\
ByT5$\dag$ & 35.9/15.8/29.8 & 32.7/12.2/27.2 & 31.4/11.0/26.2 & 35.1/13.5/29.1 &-/-/- & -/-/-  & -/-/- \\
mBART$\dag$ & 34.4/13.0/28.1 & 32.0/11.1/26.4 & 33.1/11.0/27.8
 & 38.3/15.4/32.4& -/-/- & -/-/-  & -/-/- \\
DOC-mT5$\dag$
& 37.7/16.7/31.4
& 32.4/11.9/27.0
& 33.6/12.8/28.5
& 36.8/15.0/31.5
& 34.2/13.3/27.9
 & 36.3/14.3/30.8
& 35.2/13.7/29.5
\\
Baseline Model & 42.4/19.7/36.4 & 38.5/15.8/32.9 & 34.9/13.4/29.7 & 36.9/14.8/31.4& 40.9/18.0/35.0& 37.3/14.9/31.9& 37.8/15.3/32.2\\
Baseline Model$^+$ & 42.6/19.6/36.6 & 38.8/16.1/33.1 & 34.9/13.3/29.6 & 37.1/14.8/31.5& 40.7/18.1/34.9& 37.2/14.9/31.8& 37.6/15.2/31.9\\
 \hline
\textbf{TRIP (Ours)} &
\textbf{45.3/22.5/39.0} &
\textbf{40.8/17.3/34.4} &
36.6/\textbf{14.6}/30.8 &
38.7/15.9/32.7&
 \textbf{42.8/19.9/36.8} &
\textbf{38.5/16.0/32.9} &
\textbf{39.4/16.4/33.3}
\\
\hline
\end{tabular}
\caption{\label{wiki}
Results for cross-lingual abstractive summarization on Wikilingua in (X $\rightarrow$ En). We report the scores of F-measure for ROUGE-1/ROUGE-2/ROUGE-L. -: the score is not reported. $\dag$: the scores are taken from \citet{lee-etal-2022-docmt5} and the official GEM benchmark \citep{gehrmann-etal-2021-gem}: \url{https://gem-benchmark.com/results}.
}
\end{table*}
\label{abmt}
\paragraph{Case Study}
Table \ref{case} presents three case studies that demonstrate and compare the outputs between TRIP and the baseline systems. We highlight the correct translation in aqua and the wrong translation in hot pink. In addition to the comparison to the Baseline Models, we also present the outputs from popular commercial translation systems Google Translate, Microsoft Translator, and DeepL Translate. Each case demonstrates that TRIP is the best in terms of three characteristics respectively: (i) tense consistency \citep{blonde, sun-etal-2022-rethinking} across the sentences, (ii) noun-related issues \citep{blonde} such as singular and plural consistency as well as attaching definite article `the' to a previously mentioned object `light', and (iii) conjunction presence that indicates the relationship between sentences and makes the translation natural and fluent \citep{10.1609/aaai.v33i01.33017338, sun-etal-2022-rethinking}. While some translations in the third case are acceptable, missing coordinating conjunction does not precisely capture the relationship between sentences and can make the translation less fluent.
\par
TRIP is the best one among the systems at resolving cross-lingual ambiguities. This observation highlights the necessity of translating with document-level contexts for resolving cross-lingual ambiguities. The observations also align with the BlonDe measurements reported above.

\subsection{Cross-lingual Abstractive Summarization}
\paragraph{Experimental Settings}
We follow the same setting used by DOCmT5 \citep{lee-etal-2022-docmt5} to evaluate cross-lingual abstractive summarization on the benchmark of Wikilingua \citep{ladhak-etal-2020-wikilingua}. The only difference is that they put a special prefix \textit{"Summarize X to Y"} where X and Y are the source and target language tags for summarization like mT5. We put a target language tag as the prefix. We use the F1 measure for ROUGE-1/ROUGE-2/ROUGE-L scores \citep{lin-2004-rouge} for evaluation.
\paragraph{Baseline Systems}
We report the scores for DOCmT5 taken from \citet{lee-etal-2022-docmt5}, and we use prior SOTA scores from the official GEM benchmark \citep{gehrmann-etal-2021-gem} for mT5, ByT5 \citep{xue-etal-2022-byt5}. We also employ mBART and the Baseline Models as the baselines. See Appendix \ref{npa} for the number of model parameters.
\paragraph{Results}
Table \ref{wiki} demonstrates that TRIP clearly exceeds previous SOTA systems in several directions, including up to 8.9 ROUGE-L points in (Hi $\rightarrow$ En) compared to DOCmT5. Hence, we conclude that TRIP is an effective multilingual pre-training framework for cross-lingual abstractive summarization. We postulate that the improvement is attributed to the trilingual pre-training objective overlooked by previous works such as DOCmT5.
\par
Also, we found that using bilingual data for Baseline Model$^+$ seems less beneficial on Wikilingua for cross-lingual abstractive summarization. TRIP clearly surpasses the Baseline Model$^+$. This observation indicates that the trilingual objective with the Grafting mechanism is superior to the conventional bilingual objectives for multilingual pre-training.
\paragraph{Case Study} Table \ref{case_summarization} 
 in Appendix shows three case studies that TRIP outputs better abstractive cross-lingual summarization. For space reasons, we leave more details in Appendix \ref{casesum}.
 \section{Related Work}
\subsection{Multilingual Pre-training}
Multilingual pre-training has achieved great success. Previous works can be categorized into two streams: monolingual pre-training \citep{conneau-etal-2020-unsupervised,liu-etal-2020-multilingual-denoising, xue-etal-2021-mt5} and bilingual pre-training \citep{huang-etal-2019-unicoder, chi-etal-2021-infoxlm,ouyang-etal-2021-ernie,tang-etal-2021-multilingual, 2021arXiv210408692C, reid-artetxe-2022-paradise, lee-etal-2022-docmt5}.
Monolingual pre-training uses monolingual corpora in many different languages and perturbs the inputs with sentence permutation  \citep{liu-etal-2020-multilingual-denoising} and span corruption \citep{xue-etal-2021-mt5} and requires the model to reconstruct the original input. \citet{reid-artetxe-2022-paradise} also proposes dictionary denoising on monolingual data. 
For bilingual pre-training, \citet{tang-etal-2021-multilingual} uses clean sentence-level bilingual translation pairs on pre-trained models to improve MT. \citet{2021arXiv210408692C} extends mT5 with objectives such as translation span corruption. DOCmT5 \citep{lee-etal-2022-docmt5} creates synthetic translation pairs and uses sentence permutation for a document-level multilingual pre-training.
\par
So far, document-level multilingual pre-training with parallel data is currently understudied in the field, and all the works mentioned above overlooked the value of trilingual parallel data.
\par
\subsection{Document-level Cross-lingual Tasks} 
Document-level MT and cross-lingual abstractive summarization are the two document-level cross-lingual tasks that we investigate in this paper. 
\par
Document-level MT \citep{miculicich-etal-2018-document,maruf-etal-2019-selective,10.1145/3441691,2022arXiv220913940L} is a challenging translation task, possibly due to the long input problem \citep{gtrans2,gtrans3} when directly modelling the long document and the necessity in understanding contexts \citep{voita-etal-2018-context,voita-etal-2019-good}. Therefore, many works focus on using sentence-level models with a smaller contextual window to simulate document-level MT \citep{han1, han2}. This paper follows the challenging setting \citep{bao-etal-2021-g, lee-etal-2022-docmt5} that directly optimizes a document-level model with a longer context window that provides a richer source of context, which is also a double-edged sword that could be harder due to the long input problem.
\par
Abstractive summarization is a generation task that requires an understanding of texts \citep{chopra-etal-2016-abstractive, fan-etal-2018-controllable}. We focus on a cross-lingual setting where source and target are written in different languages \citep{ladhak-etal-2020-wikilingua}.
\section{Conclusions}
We present a novel sequence-to-sequence multilingual document-level pre-training methodology called \textbf{TRIP}, which is the first in our field to propose a trilingual objective for multilingual pre-training through Grafting. We also propose a  novel method to construct high-quality trilingual document pairs. Experimental results indicate that TRIP achieves competitive SOTA scores on both multilingual document-level machine translation and cross-lingual abstractive summarization. Future works could improve TRIP to include polygonal parallel translation pairs in multilingual pre-training. We plan to release the model checkpoints and a manually annotated benchmark created using our created document-level corpus MTDD to facilitate future research on multilingual document-level MT.

\section*{Limitations}
\paragraph{TRIP} TRIP leverages high-quality document-level trilingual translation pairs for pre-training on multilingual models. It is usually harder to collect high-quality trilingual data than to collect monolingual data written in different languages used by conventional methods. While we can possibly relax the quality bar for the data, additional experiments should be done to verify this view.
\paragraph{MTDD} We create MTDD, a corpus that is composed of trilingual document pairs. It could be further extended to include polygonal parallel document pairs to provide a stronger signal for multilingual pre-training. We leave this to future works.
\paragraph{Large Language Models} Large Language Models (LLMs) such as ChatGPT have shown good translation abilities \citep{2023arXiv230506575L}, while they still lag behind supervised systems \citep{2023arXiv230108745J, 2023arXiv230404675Z}. We do not directly compare them, as they are much larger in their number of parameters than the systems described in this work. 
\section*{Ethics Statement}
We honour and support the EMNLP Code of Ethics. The datasets used in this work are well-known and widely used, and the dataset pre-processing does not use any external textual resource. We also curate a corpus for pre-training language models. Although we have made our best efforts in reducing potentially offensive and toxic data, the models are subject to generating offensive context. But the issues mentioned above are widely known to exist for these models commonly. Any content generated does not reflect the view of the authors.
\bibliography{custom}
\bibliographystyle{acl_natbib}
\newpage
\appendix
\begin{table*}[h!]
\scriptsize
\centering
\setlength\aboverulesep{0pt}\setlength\belowrulesep{0pt}
\setcellgapes{1pt}\makegapedcells
    \setlength\tabcolsep{5pt}
    \setlength\extrarowheight{2pt}
\begin{tabular}{l|p{11.5cm}}
\hline
\multicolumn{2}{c}{\textit{Case 1}}
\\
\hline
\textbf{Source} & Ayakkabılarını (ve bağcıklarınla tabanlıklarını) kuruması için orta derecede ışık alan bir yere koy. Sıcak bir yere (örneğin, radyatörün yanına) ya da doğrudan güneş ışığına koyma çünkü bu, ayakkabılara zarar verebilir. Ayakkabılarını kurutucuya koymak tavsiye edilmez çünkü kurutucu, ayakkabı tabanlarını yamultabilir.\\
\hline
\textbf{Source (Google-translated)}& Put your shoes (and your laces and insoles) in a moderately light place to dry. Do not place it in a hot place (for example, near a radiator) or in direct sunlight as this may damage the shoes. Putting your shoes in the dryer is not recommended because the dryer can warp the soles of your shoes.\\
\hline
\textbf{Reference} & Air-dry your shoes.\\
\hline
\textbf{Baseline Model (Document-level)} & Allow your shoes (and laces) to dry. \\
\hline
\hline
\textbf{TRIP} & Let your shoes (and the laces) air dry.\\
\hline
\hline
\multicolumn{2}{c}{\textit{Case 2}}
\\
\hline
\textbf{Source} & Bunun için yeşil bir arka plan üzerindeki beyaz konuşma balonuna dokun. Ana Ekranlarından birinde olması gerekir. ’ye dokun. Mesajlar ekranının sol üst köşesindedir. Açık bir sohbetin varsa Mesajlar menüsüne dönmek için ekranın sol üst köşesindeki < butonuna dokun. ’e dokun. Ekranının sağ alt köşesindedir. Seçilen mesajların silinir.\\
\hline
\textbf{Source (Google-translated)} & To do this, tap the white speech bubble on a green background. It should be on one of their Home Screens. Tap It's in the upper-left corner of the Messages screen. If you have an open chat, tap the < button in the upper left corner of the screen to return to the Messages menu. Tap . It's in the lower right corner of your screen. The selected messages are deleted.\\
\hline
\textbf{Reference} & Open your iPhone's messages. Tap Edit. Select each conversation you wish to delete. Tap Delete. \\
\hline
\textbf{Baseline Model (Document-level)} & Open your iPhone's Settings . Tap Messages. Tap Delete Messages.\\
\hline
\hline
\textbf{TRIP} & Open Messages. Tap the Messages tab. Tap Delete. Tap Delete to confirm.\\
\hline
\hline
\multicolumn{2}{c}{\textit{Case 3}}
\\
\hline
\textbf{Source} & Bazıları için geçmiş yaşamlar gidilecek bir yer değil, seni sen yapan şeyin bir kısmıdır. İnsanlığın tarihi boyunca birçok kültür reenkarnasyonu inançlarının merkezine koymuştur. İslam ve Hıristiyanlık reenkarnasyona inanmasa da, Hinduistler, bazı Museviler ve bazı Budistler buna inanır. En iyisi kendini bir dine tümüyle adamaktansa (çünkü dinler çok kısıtlayıcı olabilir) kendi yolunu keşfetmen. Kendi manevi doğrularını kendin bul.\\
\hline
\textbf{Source (Google-translated)} & For some, past lives are not a place to go but part of what makes you who you are. Throughout the history of humanity, many cultures have put reincarnation at the center of their beliefs. Although Islam and Christianity do not believe in reincarnation, Hinduists, some Jews, and some Buddhists do. It's best to explore your own path rather than devote yourself entirely to a religion (because religions can be too restrictive). Find your own spiritual truth.\\
\hline
\textbf{Reference} & Become spiritual.\\
\hline
\textbf{Baseline Model (Document-level)} & Understand that some people believe in reincarnation. Find your own way. \\
\hline
\hline
\textbf{TRIP} & Explore your own spiritual journey.
\\
\hline
\end{tabular}
\caption{\label{case_summarization}
Three case studies from Wikilingua (Tr $\rightarrow$ En) demonstrate that TRIP outputs better summarization.
}
\end{table*}
\clearpage
\section{Unseen (X$\rightarrow$X) Language Pairs on MT}
\label{unseenpair}
\begin{figure}[t!]
\begin{center}
\centerline{
\includegraphics[width=7cm]{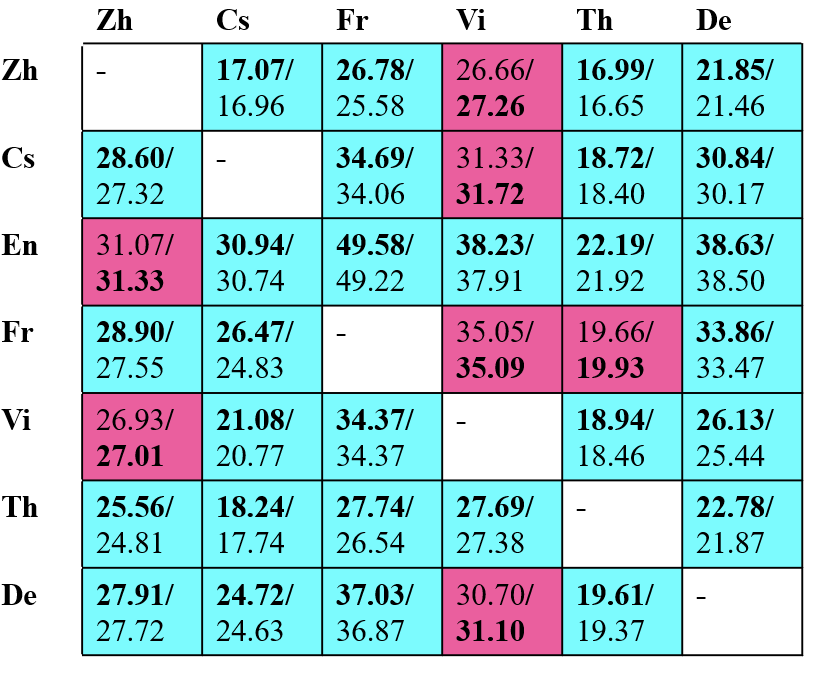}}
\caption{Results on TED Talks in (X$\rightarrow$X) with our TRIP checkpoint pre-trained in (X$\rightarrow$En) directions only. The scores are written in TRIP as the former and the Baseline Model as the latter. Rows represent the source languages and columns represent the target languages. We highlight in aqua when TRIP wins (darker one when printed in B\&W) and in hot pink (lighter one when printed in B\&W) when the Baseline Model wins.
}
\label{iwslt_xx}
\end{center}
\end{figure}

Figure \ref{iwslt_xx} reports the performance on TED Talks in the direction of (X$\rightarrow$X) with our TRIP checkpoint pre-trained in (X$\rightarrow$En) directions with our corpus. The row represents the translation source language and the column represents the translation target language. TRIP clearly improves most of these translation directions which are unseen during pre-training. This indicates that fact the TRIP can generalize the cross-lingual understanding ability to unseen language pairs. This aligns with the fact reported in \citet{lee-etal-2022-docmt5}.
\section{Case Study on Summarization}
\label{casesum}
Table \ref{case_summarization} 
 shows that TRIP outputs better summarization in (i) precisely capturing the context in Case 1, (ii) outputting consistent nouns, i.e., `messages' instead of `settings' in Case 2 and (iii) producing concise and accurate summarization in Case 3. This highlights that TRIP captures better cross-lingual understanding than the baseline system, which effectively mitigates cross-lingual ambiguities.
\section{Number of Model Parameters}
\label{npa}
\newpage
\begin{table}[htb]
\centering
\setlength\tabcolsep{7pt}
\setlength\aboverulesep{0pt}\setlength\belowrulesep{0pt}
\setcellgapes{3pt}\makegapedcells
\begin{tabular}{l|c}
\hline
\textbf{Model} & \textbf{Number of Parameters}\\
\hline
M2M-100 & 418M\\
mBART & 611M\\
MARGE & 963M\\
mT5 & 1.23B$^*$\\
DOCmT5 & 1.23B$^*$\\
ByT5-Small & 300M\\
ByT5-Base & 582M\\
ByT5-Large & 1.23B$^*$\\
mT5-XL & 3.74B\\
Baseline Model & 862M\\
Baseline Model$^+$ & 862M\\
\hline
\textbf{TRIP (Ours)} & 862M\\

\hline
\end{tabular}
\caption{\label{np}
Comparison in the number of parameters for the pre-trained models used in our experiments. $*$: these models all use the model architecture of mT5-Large, and we report the number of model parameters taken from the original paper of mT5 reported by \citet{xue-etal-2021-mt5}.
}
\end{table}
\begin{table}[htb!]
\scriptsize
\centering
\setlength\tabcolsep{7pt}
\setlength\aboverulesep{0pt}\setlength\belowrulesep{0pt}
\setcellgapes{2pt}\makegapedcells
\begin{tabular}{ccc|ccc}
\hline
\textbf{Source} & \textbf{Target} & \textbf{Size/GB} & \textbf{Source} & \textbf{Target} & \textbf{Size/GB}  \\
\hline
\textbf{Es} & \textbf{En} & 3.22 &  \textbf{Pt} & \textbf{Es} & 2.71 \\
\textbf{Es} & \textbf{Ca} & 2.07 & \textbf{Uk} & \textbf{Ru}  & 1.60\\
\textbf{Fr} & \textbf{Es} & 1.47 & \textbf{Es} & \textbf{Pt} & 1.47\\
\textbf{En} & \textbf{De} & 1.25 & \textbf{Pt} & \textbf{En} & 1.14\\
\textbf{Ca} & \textbf{Es} & 1.12 & \textbf{Fr} & \textbf{En} & 1.03 \\
\textbf{Ru} & \textbf{Uk} & 0.87& \textbf{Pt} & \textbf{Fr} & 0.73\\
\hline
\end{tabular}
\caption{\label{mind_lang}
A language list in ISO code for the top 12 language directions for the bilingual high-quality pre-training data to illustrate the scale of size.
}
\end{table}
Table \ref{np} presents the number of model parameters for the pre-trained models used in our experiments.
\par
For the scores of ByT5 presented in Table \ref{wiki}, we report the maximum scores for each direction among ByT5-Base, ByT5-Small, and ByT5-Large. This is due to space reasons. See \url{https://gem-benchmark.com/results} for the tailored scores.
\section{MTDD Corpus Scale}
\label{mtddscale}
Table \ref{mind_lang} presents the top-12 English-centric bilingual data statistics to illustrate the scale of MTDD. The total size of the data is about 40/80 GB respectively for the bilingual and the trilingual data applied with Grafting.
\end{document}